%% file: main_jp.tex
\documentclass[conference]{IEEEtran}
\IEEEoverridecommandlockouts

\usepackage{cite}
\usepackage{amsmath,amssymb,amsfonts}
\usepackage{algorithm}
\usepackage{algpseudocode}
\usepackage{graphicx}
\usepackage{textcomp}
\usepackage{xcolor}
\usepackage{booktabs}
\usepackage{multirow}
\usepackage{url}
\usepackage{float}
\usepackage{dsfont}
\usepackage{hyperref}
\hypersetup{colorlinks=true,allcolors=blue}

\def\BibTeX{{\rm B\kern-.05em{\sc i\kern-.025em b}\kern-.08em
    T\kern-.1667em\lower.7ex\hbox{E}\kern-.125emX}}

\begin{document}

\title{Bayesian Spectral Emotion Transition Discovery from Multi-Annotator Disagreement}


\author{
\IEEEauthorblockN{Keito Inoshita\textsuperscript{1,2,*} \quad Takato Ueno\textsuperscript{3}}
\IEEEauthorblockA{
\textsuperscript{1}Faculty of Business and Commerce, Kansai University, Suita, Japan\\
\textsuperscript{2}Data Science and AI Innovation Research Promotion Center, Shiga University, Hikone, Japan\\
\textsuperscript{3}Graduate School of Data Science, Shiga University, Hikone, Japan\\
inosita.2865@gmail.com \quad s7025101@st.shiga-u.ac.jp
}
}

\maketitle

\input{sections/00_abstract}
\input{sections/01_introduction}
\input{sections/02_related_work}
\input{sections/03_methods}
\input{sections/04_experiments}
\input{sections/05_discussion}
\input{sections/06_conclusion}

\bibliographystyle{IEEEtran}
\bibliography{references_bsetd}

\input{sections/A_appendix}

\end{document}

%% file: sections/00_abstract.tex
\begin{abstract}
Emotions evolve through the dynamics of conversation, and understanding their transition structure is foundational to applications ranging from mental-health screening to dialogue systems. However, existing studies typically compress multi-rater judgments into a single hard label by majority voting, discarding the uncertainty signal needed to understand turn-to-turn transitions. In this article, we propose Bayesian Spectral Emotion Transition Discovery (BSETD), a two-stage framework that discovers emotion-transition structure from multi-rater soft labels. In the first stage, a hierarchical Dirichlet--Multinomial posterior is constructed through the outer product of soft labels, equipping each cell of the $K \times K$ transition matrix with a credible interval and Benjamini--Hochberg (BH) false discovery rate (FDR)-controlled significance. In the second stage, the symmetrized graph Laplacian is spectrally decomposed to separate a low-frequency (inertia) component from a high-frequency (contagion) component. On EmotionLines, BSETD simultaneously recovers the signatures of two distinct affective spaces: the Plutchik-adjacent transitions disgust$\to$anger ($\log_2$ lift $+0.94$) and anger$\to$disgust ($+0.86$) are over-represented, while the Russell-valence-reversed transitions joy$\to$anger ($-0.90$) and anger$\to$joy ($-0.89$) are under-represented. A five-source cross-corpus validation yields pairwise Pearson correlations in $0.91$--$0.98$ within English, $0.79$--$0.85$ against Chinese M3ED, and $0.979$ between the human hard labels and the LLM virtual soft labels on the same utterance set, demonstrating that a pipeline preserving annotator uncertainty bridges the computational study of emotion dynamics with established psychological theory.

\end{abstract}


%% file: sections/01_introduction.tex
\section{Introduction}
\label{sec:introduction}

Emotions are not static attributes of utterances but rather evolve through the dynamics of human conversation. A speaker who begins a dialogue in a neutral mood may shift to joy in response to encouragement from another party, sustain that joy for several turns, and abruptly transition to surprise upon receiving unexpected news. Such temporal patterns describe the structure of emotional dynamics in conversation: how one emotion follows another, how an emotion persists within a single speaker, and how it propagates between dialogue partners~\cite{kuppens2017inertia}. Understanding this structure is foundational to applications ranging from mental-health screening to socially intelligent human--machine dialogue~\cite{picard1997affective}.

Despite the importance of emotion dynamics, prior studies have systematically discarded a critical source of information, namely, annotator disagreement. The standard practice in annotating conversational emotion corpora is to compress multi-rater judgments into a single hard label by majority voting. In EmotionLines~\cite{hsu2018emotionlines}, 72.6\% of utterances exhibit disagreement among the five raters and 14.3\% fail to receive a majority vote, figures that reflect genuine perceptual ambiguity rather than annotator error. Forcing such utterances into a single category irreversibly destroys the uncertainty signal embedded in the vote distribution. Recent work has shown that large language models (LLM) cannot reproduce the shape of human uncertainty distributions~\cite{inoshita2026llm}, and the perspectivist paradigm argues that disagreement should be treated as meaningful signal rather than removable noise~\cite{perspectivist2021}. A method that exploits the full multi-rater vote distribution -- that is, soft labels -- thus has the potential to recover richer and more reliable transition patterns than any existing approach.

Existing studies on Emotion Recognition in Conversation (ERC) have demonstrated substantial progress in predicting next-turn emotion labels~\cite{dialoguernn2019,emotionIC2024,graphsmile2023,dgode2024}. However, these methods share two fundamental limitations. First, they treat emotion labels as deterministic targets to be predicted rather than as distributions to be understood, leaving the structural question of how emotions transition unaddressed. Second, uncertainty is only implicitly quantified through prediction confidence, and per-annotator disagreement is never explicitly modeled. Parallel work on relational discovery in dialogue targets intent drift~\cite{wang2025intentdrift} or emotion-grounded utterance generation~\cite{mecot2025} rather than structural discovery. As a result, no existing method jointly addresses (i) estimating a probabilistic transition matrix from soft multi-rater labels, (ii) decomposing it into interpretable frequency-domain components, and (iii) cross-validating the discovered patterns against psychological theory across multiple corpora.

In this study, we propose Bayesian Spectral Emotion Transition Discovery (BSETD), a two-stage framework for the discovery of statistically significant emotion transition patterns in multi-annotator dialogue corpora. In the first stage, a hierarchical Dirichlet--Multinomial (DM) model is constructed, in which the uncertainty of each utterance pair is encoded into the transition count matrix via the outer product of soft labels; the posterior is obtained in closed form through DM conjugacy, and the Benjamini--Hochberg (BH) procedure~\cite{benjamini1995fdr} controls the false discovery rate (FDR) at level $q^{*}$ over the $j \to k$ entries. In the second stage, the estimated transition matrix is treated as a weighted graph adjacency matrix, and a spectral decomposition based on the symmetrized graph Laplacian~\cite{chung2005laplacians} is applied. As empirically verified in Section~\ref{sec:exp:main4}, low-eigenvalue components align with the emotional inertia of Kuppens \& Verduyn~\cite{kuppens2017inertia}, whereas high-eigenvalue components align with the emotional contagion of Hatfield \emph{et al.}~\cite{hatfield1993contagion}, providing a mathematically principled bridge between the Bayesian posterior and these psychological constructs.

The main contributions of this article are summarized as follows.
\begin{itemize}
  \item[i)] A hierarchical DM posterior estimator based on soft-label outer products is newly designed, in which annotator uncertainty is propagated to transition-probability estimation with 95\% credible intervals and BH-FDR-controlled significance.
  \item[ii)] An inertia--contagion spectral decomposition based on the symmetrized graph Laplacian is constructed, in which the estimated transition matrix is separated into low-frequency and high-frequency components corresponding to the constructs of Kuppens~\cite{kuppens2017inertia} and Hatfield~\cite{hatfield1993contagion}.
  \item[iii)] A five-source cross-corpus evaluation is developed over EmotionLines, MELD, DailyDialog, the Chinese M3ED, and a GPT-5.4-mini virtual-annotator variant, achieving pairwise Pearson correlations in $[0.79, 0.98]$ and computationally verifying Plutchik adjacency, the Gottman cascade, and Hatfield contagion.
\end{itemize}

The rest of this paper is organized as follows. Section~\ref{sec:related} reviews related work, and Section~\ref{sec:method} elaborates the proposed BSETD framework. Section~\ref{sec:experiments} presents the experimental results and ablation studies, while Section~\ref{sec:discussion} discusses the main findings and limitations. Finally, Section~\ref{sec:conclusion} concludes this article and outlines directions for future research.

%% file: sections/02_related_work.tex
\section{Related Work}
\label{sec:related}

\subsection{Emotion Dynamics and Recognition in Conversation}
\label{sec:related:erc}
ERC has evolved along two axes: recurrent architectures that model the temporal flow of emotional states, and graph-based architectures that capture relational dependencies. DialogueRNN~\cite{dialoguernn2019} incorporates dedicated modules for global context, speaker state, and emotion state into a gated recurrent architecture as a widely adopted baseline. EmotionIC~\cite{emotionIC2024} extends this line by decomposing emotional influence into an inertia component (speaker's history) and a contagion component (influence from interlocutors).

EmotionIC is terminologically the closest to BSETD and merits a precise comparison along three axes. In representation, EmotionIC encodes inertia and contagion as implicit internal states of a recurrent classifier, whereas the BSETD counterparts are the diagonal and off-diagonal blocks of a $K \times K$ posterior with closed-form credible intervals. In objective, EmotionIC is trained with cross-entropy against majority-vote labels and reports only classification accuracy, whereas BSETD is an unsupervised Bayesian estimator whose output is the transition structure itself with BH-FDR-controlled significance. In task, EmotionIC targets next-turn label prediction, whereas BSETD targets corpus-level structural discovery.

GraphSmile~\cite{graphsmile2023} and DGODE~\cite{dgode2024} represent the graph-based line, constructing a dynamic interaction graph over dialogue turns and using graph neural propagation to capture cross-speaker influence. While achieving strong performance on IEMOCAP~\cite{iemocap2008} and MELD~\cite{poria2019meld}, they share the same stance as the rest of the prior literature -- next-label prediction rather than statistical-structure discovery -- and none model annotator uncertainty or produce credible intervals. Differing from all of the above, BSETD is designed not as a predictive system but as a discovery framework: it estimates the posterior over the full $K \times K$ transition matrix and identifies which $j \to k$ pairs are statistically distinguishable from a uniform baseline.

\subsection{Soft Labels and Annotator-Disagreement Modeling}
\label{sec:related:softlabel}
A growing line of research argues that annotator disagreement in subjective natural language processing tasks should be preserved and modeled rather than resolved into a single ground-truth label. Cowen and Keltner~\cite{cowen2017emotion} demonstrated that human emotion experience is best described by a high-dimensional continuous space rather than by discrete categories, implying that forced-choice categorical labels necessarily discard perceptual information. The Learning-with-Disagreements (LeWiDi) shared task~\cite{lewidi2025} institutionalized this view in NLP by evaluating systems on their ability to reproduce the full annotator distribution rather than the majority vote. The perspectivist paradigm~\cite{perspectivist2021} further argues that aggregating diverse annotator perspectives into a single label actively suppresses the information needed to understand subjective phenomena.

Within this general trend, constraints specific to the affective domain have also been identified. Inoshita \emph{et al.}~\cite{inoshita2026llm} systematically compared human and LLM annotations across multiple emotion corpora and showed that LLMs reliably capture the modal label yet fail to reproduce the shape of human uncertainty distributions. This finding implies that human disagreement signals cannot be substituted by LLM annotations, and that pipelines exploiting genuine multi-rater soft labels are needed. Stage 1 of BSETD addresses this requirement: through the soft-label outer-product formulation, the annotator distribution is propagated from the utterance level to the transition-probability level in a statistically consistent manner.

\subsection{Relational Discovery and Graph-Spectral Methods in Dialogue}
\label{sec:related:discovery}
Recent work that seeks to discover relational structure in dialogue data beyond emotion classification has been proposed, but all of these target a different objective than the present study. Wang \emph{et al.}~\cite{wang2025intentdrift} apply temporal transition accumulation to change-point inference of intent drift within a single conversation; MECoT~\cite{mecot2025} addresses the generative quality of role-playing through a Markov emotional chain-of-thought; and the Knowledge-Bridged Causal Interaction Network (KBCIN)~\cite{kbcin_aaai2023} pursues predictive causal emotion entailment under hard-label inputs. Differing from these, BSETD infers, at the corpus level and under BH-FDR control, which pairwise transitions exist directly from soft labels. The Affective Ising Model (AIM) of Loossens \emph{et al.}~\cite{loossens2020affective} is the closest theoretical precedent: inspired by statistical mechanics, it models the moment-to-moment fluctuation of affect as a stochastic process on a free-energy landscape, capturing inertia as a tendency toward a local minimum and instability as a reversal between homebases. However, AIM is formulated in a two-dimensional positive--negative valence space, and extending it to the seven Ekman categories (42 directed pairs) requires the multi-category Bayesian posterior of Stage 1.

As a methodological foundation, graph-spectral methods have also seen increasing application to affective computing. GS-MCC~\cite{gssmcc_aaai2025} applies spectral filtering to the inter-utterance graph in ERC and separates speaker-level signals (low frequency) from cross-speaker signals (high frequency). BSETD adopts the same spectral-decomposition idea but applies it to a graph over emotion categories rather than over utterances -- a $K \times K$ graph whose edge weights are Bayesian posterior transition probabilities. As a result, the output is directly interpretable as the psychological constructs of inertia and contagion rather than as latent speaker embeddings. The theoretical basis of the Stage 2 decomposition follows Chung's graph-Laplacian framework~\cite{chung2005laplacians}.

In summary, as shown in Table~\ref{tab:comparison}, existing relational-discovery methods are constrained to intent, generation, binary affect, or hard labels, and existing spectral methods remain at the utterance-level graph. BSETD jointly addresses soft-label multi-category transition estimation, category-level spectral decomposition, and multi-corpus validation within a single framework.

\begin{table}[t]
  \centering
  \caption{Comparison of closely related methods and BSETD across five axes: soft-label input, uncertainty quantification, discovery objective, spectral decomposition, and connection to psychological theory. $\checkmark$ = satisfied, ``partial'' = implicit or limited, blank = absent.}
  \label{tab:comparison}
  \footnotesize
  \setlength{\tabcolsep}{3pt}
  \begin{tabular}{lccccc}
    \hline
    Method & Soft & Uncert.\ & Discovery & Spectral & Psych. \\
    \hline
    DialogueRNN~\cite{dialoguernn2019}        &            &             &           &          &                   \\
    EmotionIC~\cite{emotionIC2024}            &            &             &           &          & partial           \\
    GraphSmile~\cite{graphsmile2023}          &            &             &           &          &                   \\
    DGODE~\cite{dgode2024}                    &            & partial     &           &          &                   \\
    GS-MCC~\cite{gssmcc_aaai2025}             &            &             &           & \checkmark &                 \\
    KBCIN~\cite{kbcin_aaai2023}               &            &             &           &          &                   \\
    LeWiDi-2025~\cite{lewidi2025}             & \checkmark & partial     &           &          &                   \\
    AIM~\cite{loossens2020affective}          &            & partial     & \checkmark &          & \checkmark       \\
    \hline
    \textbf{BSETD (ours)}                     & \checkmark & \checkmark  & \checkmark & \checkmark & \checkmark      \\
    \hline
  \end{tabular}
\end{table}

%% file: sections/03_methods.tex
\section{BSETD Framework for Emotion Transition Discovery}
\label{sec:method}

\begin{figure*}[t]
  \centering
  \includegraphics[width=0.75\linewidth]{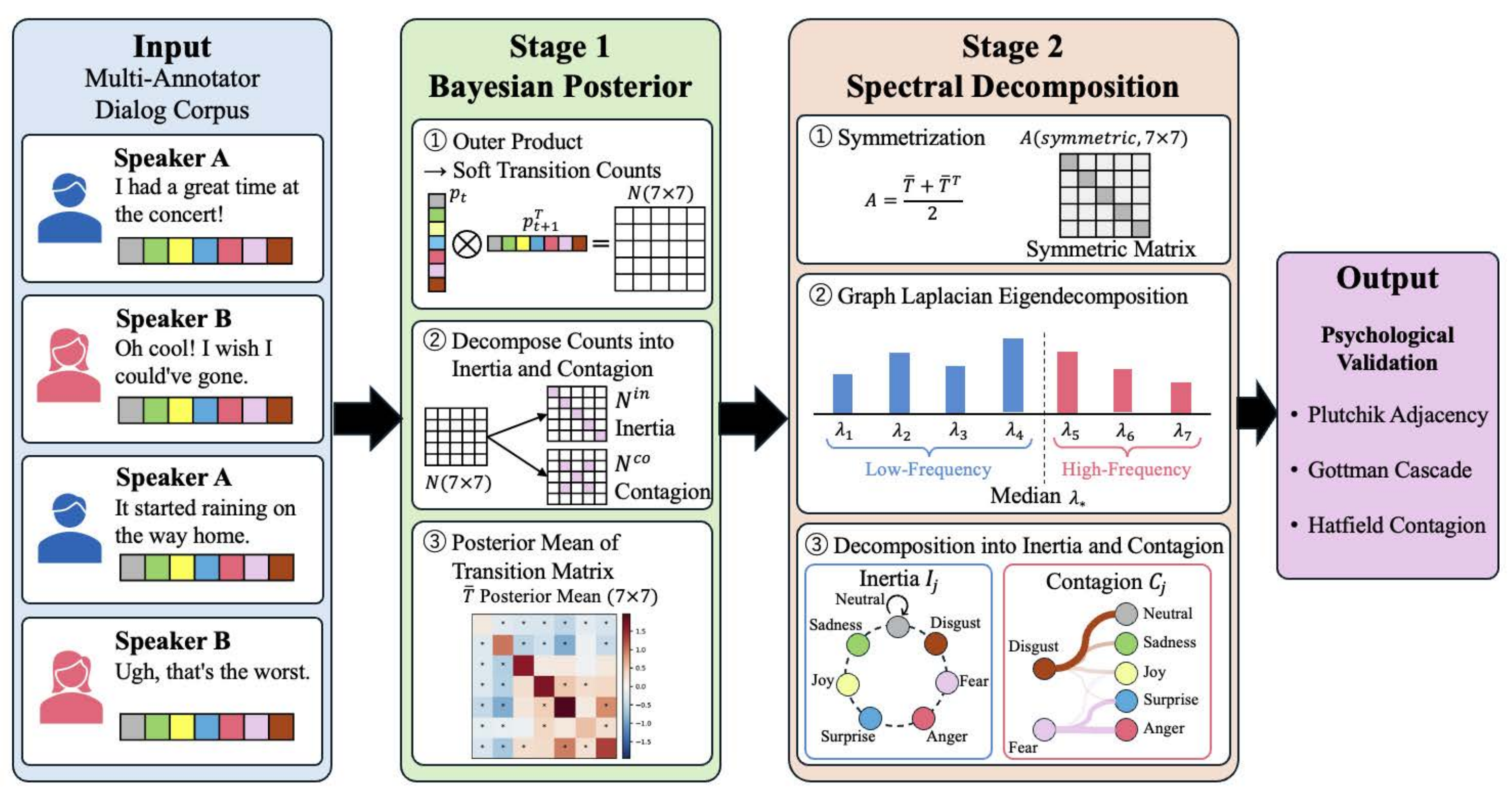}
  \caption{Overview of the BSETD Pipeline. Stage 1 constructs the hierarchical DM posterior over the $K \times K$ transition matrix from multi-rater soft labels via outer-product accumulation, and splits the counts into intra-speaker (inertia) and inter-speaker (contagion) sub-matrices. Stage 2 symmetrizes the posterior-mean transition matrix and applies graph Laplacian spectral decomposition, yielding inertia indices $I_j$ from the low-frequency band and contagion indices $C_j$ from the high-frequency band.}
  \label{fig:pipeline}
\end{figure*}

\subsection{Problem Definition}
\label{sec:method:setup}

BSETD consists of two sequential stages, illustrated in Fig.~\ref{fig:pipeline}: Stage 1 estimates the posterior of a $K \times K$ emotion-transition matrix under a hierarchical DM model with empirical-Bayes (EB) concentration estimation, and Stage 2 symmetrizes that matrix following Chung~\cite{chung2005laplacians} and applies the spectral decomposition of the normalized Laplacian to separate a low-frequency (inertia) and a high-frequency (contagion) component. A conversational corpus $\mathcal{D} = \{d_1, \ldots, d_N\}$ consists of $N$ dialogues, where each dialogue $d$ is a temporally ordered sequence of utterances $u_{d,1}, u_{d,2}, \ldots$. Each utterance $u_{d,t}$ is associated with a soft label $\mathbf{p}_{d,t} \in \Delta^{K-1}$, a probability vector over $K$ emotion categories, constructed either directly from the votes of $R$ independent raters or, in the absence of human votes, from a downstream surrogate distribution. In this article, we adopt the $K = 7$ Ekman category set (neutral, joy, sadness, fear, anger, surprise, disgust). Each utterance is also annotated with a speaker identifier $s_{d,t}$, enabling the separation of intra-speaker and inter-speaker dynamics.

The structural question targeted by BSETD is formulated as follows. Let $\mathbf{T}$ denote a latent corpus-level $K \times K$ transition matrix whose entry $T_{jk}$ is the conditional probability that an utterance labeled with emotion $j$ is followed within the same dialogue by an utterance labeled with emotion $k$. Owing to the soft labels and the finite-sample regime, $\mathbf{T}$ is a posterior quantity rather than a point estimate. We seek (i) a calibrated posterior $p(\mathbf{T} \mid \mathcal{D})$ that respects annotator uncertainty, (ii) tests of which $j \to k$ entries deviate significantly from the marginal-independence baseline, and (iii) a structural decomposition that aligns with the psychological constructs of inertia and contagion.

\subsection{Stage 1: Hierarchical DM Posterior}
\label{sec:method:stage1}

\paragraph{Soft-label transition counts.}
For an adjacent utterance pair $u_{d,t}, u_{d,t+1}$ with soft labels $\mathbf{p}_{d,t}, \mathbf{p}_{d,t+1}$, the soft transition contribution is defined as the outer product $\mathbf{p}_{d,t} \otimes \mathbf{p}_{d,t+1}$. The $(j, k)$ entry $p_{d,t}(j) \cdot p_{d,t+1}(k)$ represents the probability mass that this utterance pair contributes to the $j \to k$ transition. Summing over all adjacent pairs within all dialogues yields
\begin{equation}
  \label{eq:soft-counts}
  N_{jk} \;=\; \sum_{d=1}^{N} \sum_{t=1}^{\lvert d\rvert - 1} p_{d,t}(j)\, p_{d,t+1}(k).
\end{equation}
This construction reduces to the standard hard-label transition counts when each $\mathbf{p}_{d,t}$ is one-hot. Splitting the inner sum by the speaker-identity indicator $\mathds{1}\{s_{d,t} = s_{d,t+1}\}$ yields the inertia matrix $N^{\text{in}}_{jk}$ (same speaker) and the contagion matrix $N^{\text{co}}_{jk}$ (different speakers), with $N = N^{\text{in}} + N^{\text{co}}$.

\paragraph{Dirichlet--Multinomial generative model.}
The $j$-th row of $\mathbf{N}$ is regarded as a soft-count realization of a Multinomial distribution parameterized by the row-conditional probability vector $\mathbf{T}_j$ over $K$ targets. A Dirichlet prior with a shared concentration vector $\boldsymbol{\alpha} \in \mathbb{R}_{>0}^{K}$ is placed on each row:
\begin{equation}
  \label{eq:dm-model}
  \mathbf{T}_j \sim \mathrm{Dir}(\boldsymbol{\alpha}), \qquad
  \mathbf{N}_{j,\cdot} \mid \mathbf{T}_j \sim \mathrm{Multinomial}(N_{j,\cdot}, \mathbf{T}_j).
\end{equation}
Sharing $\boldsymbol{\alpha}$ across all rows is crucial for tying the posterior estimates of different source emotion categories together, so that the posteriors of low-mass rows are not dominated by the prior.

\paragraph{EB concentration estimation.}
The shared $\boldsymbol{\alpha}$ is obtained by Minka's fixed-point iteration~\cite{minka2000dirichlet}, which maximizes the DM marginal likelihood in closed form. With $S_j = \sum_k N_{jk}$ and $s = \sum_k \alpha_k$, the update rule is
\begin{equation}
  \label{eq:eb-update}
  \alpha_k \;\leftarrow\; \alpha_k \cdot \frac{\sum_{j} \psi(N_{jk} + \alpha_k) - \psi(\alpha_k)}{\sum_{j} \psi(S_j + s) - \psi(s)},
\end{equation}
where $\psi$ denotes the digamma function. Convergence is declared when $\max_k |\Delta \alpha_k| < 10^{-5}$. The complete Stage 1 procedure is summarized in Algorithm~\ref{alg:stage1}.

\begin{algorithm}[t]
\caption{Stage 1: Hierarchical DM Posterior.}
\label{alg:stage1}
\begin{algorithmic}[1]
\Require Dialogue corpus $\mathcal{D}$, BH-FDR level $q$, EB convergence tolerance $\epsilon$.
\State $\mathbf{N}, \mathbf{N}^{\mathrm{in}}, \mathbf{N}^{\mathrm{co}} \gets \mathbf{0}$
\For{each adjacent pair $(\mathbf{p}_{d,t}, \mathbf{p}_{d,t+1})$}
    \State $\mathbf{N} \mathrel{+}= \mathbf{p}_{d,t}\mathbf{p}_{d,t+1}^\top$
    \If{$s_{d,t} = s_{d,t+1}$}
        \State $\mathbf{N}^{\mathrm{in}} \mathrel{+}= \mathbf{p}_{d,t}\mathbf{p}_{d,t+1}^\top$
    \Else
        \State $\mathbf{N}^{\mathrm{co}} \mathrel{+}= \mathbf{p}_{d,t}\mathbf{p}_{d,t+1}^\top$
    \EndIf
\EndFor
\State Initialize $\alpha_k \gets \sum_j N_{jk} / \sum_{j,k} N_{jk}$
\Repeat
    \State Update $\alpha_k$ via Eq.~\eqref{eq:eb-update}
\Until{$\max_k |\Delta \alpha_k| < \epsilon$}
\State Compute $\bar T_{jk}, \mathrm{HDI}, \mathrm{lift}_{jk}$ from $\mathrm{Dir}(\alpha^\star + \mathbf{N}_{j,\cdot})$
\State Apply BH-FDR at level $q$ to off-diagonal cells
\State \textbf{return} $\bar{\mathbf{T}}$, HDI, $\mathrm{lift}$, BH-FDR mask, inertia/contagion sub-fits
\end{algorithmic}
\end{algorithm}

\paragraph{Posterior summary.}
Given the converged $\boldsymbol{\alpha}^\star$, the row-wise posterior is $\mathbf{T}_j \mid \mathcal{D} \sim \mathrm{Dir}(\boldsymbol{\alpha}^\star + \mathbf{N}_{j,\cdot})$. We report the posterior mean $\bar{T}_{jk}$, the 95\% Highest Density Interval (HDI) of each cell from its marginal Beta distribution, and the lift relative to the marginal distribution of the next utterance $\bar{P}_k$:
\begin{equation}
  \label{eq:lift}
  \mathrm{lift}_{jk} \;=\; \frac{\bar{T}_{jk}}{\bar{P}_k}, \qquad
  \bar{P}_k \;=\; \frac{\sum_j (N_{jk} + \alpha^\star_k)}{\sum_{j,k} (N_{jk} + \alpha^\star_k)}.
\end{equation}
The lift quantifies the multiplicative effect of conditioning on the source emotion $j$ relative to the source-independent baseline. We further report the two-sided posterior tail probability for the null $T_{jk} = \bar{P}_k$ and the BH-FDR correction~\cite{benjamini1995fdr} over the 42 off-diagonal cells. On the corpora considered in this article, the total soft-count mass is sufficiently large that the significance tests reject at any reasonable level for essentially all cells; we therefore use $\log_2(\mathrm{lift}_{jk})$ as the primary effect-size measure for ranking and reporting transitions.

\subsection{Stage 2: Symmetrized Spectral Decomposition}
\label{sec:method:stage2}

\paragraph{Adjacency matrix and Laplacian.}
The posterior-mean transition matrix $\bar{\mathbf{T}}$ is interpreted as a weighted directed graph over the $K$ emotion categories. Following Chung's directed-graph Laplacian framework~\cite{chung2005laplacians}, the symmetrization
\begin{equation}
  \label{eq:sym}
  \mathbf{A} \;=\; \tfrac{1}{2}\bigl(\bar{\mathbf{T}} + \bar{\mathbf{T}}^\top\bigr)
\end{equation}
is adopted, and the normalized Laplacian is defined as
\begin{equation}
  \label{eq:laplacian}
  \mathbf{L} \;=\; \mathbf{I} - \mathbf{D}^{-1/2} \mathbf{A}\, \mathbf{D}^{-1/2},
  \qquad
  D_{jj} = \sum_k A_{jk}.
\end{equation}
Symmetrization discards the directional component of the transition matrix while preserving the structural mass. The directional information is partially recovered by separately decomposing the inertia and contagion sub-matrices, which represent intra-speaker and inter-speaker dynamics.

\paragraph{Spectral decomposition and band-limited reconstruction.}
$\mathbf{L}$ is symmetric positive semi-definite and admits the eigen-decomposition $\mathbf{L} = \mathbf{U} \boldsymbol{\Lambda} \mathbf{U}^\top$, with eigenvalues satisfying $0 = \lambda_1 \le \lambda_2 \le \cdots \le \lambda_K \le 2$. Eigenvectors $\mathbf{u}_i$ associated with small $\lambda_i$ correspond to smooth graph signals~\cite{shuman2013emerging}, while eigenvectors with large $\lambda_i$ correspond to oscillatory signals. Splitting the spectrum at its median and letting $\mathbf{P}_{\mathrm{lo}} = \mathbf{U}_{\mathrm{lo}} \mathbf{U}_{\mathrm{lo}}^\top$ and $\mathbf{P}_{\mathrm{hi}} = \mathbf{U}_{\mathrm{hi}} \mathbf{U}_{\mathrm{hi}}^\top$ denote the orthogonal projectors onto the corresponding eigen-subspaces, the standard band-limited reconstructions on graphs~\cite{shuman2013emerging} are defined as
\begin{equation}
  \label{eq:band}
  \mathbf{A}^{\mathrm{lo}} \;=\; \mathbf{P}_{\mathrm{lo}} \mathbf{A} \mathbf{P}_{\mathrm{lo}}, \qquad
  \mathbf{A}^{\mathrm{hi}} \;=\; \mathbf{P}_{\mathrm{hi}} \mathbf{A} \mathbf{P}_{\mathrm{hi}},
\end{equation}
which are themselves symmetric by construction.
The Stage 2 procedure is summarized in Algorithm~\ref{alg:stage2}.

\begin{algorithm}[t]
\caption{Stage 2: Symmetrized Spectral Decomposition.}
\label{alg:stage2}
\begin{algorithmic}[1]
\Require Posterior-mean transition matrix $\bar{\mathbf{T}}$.
\State $\mathbf{A} \gets \tfrac{1}{2}(\bar{\mathbf{T}} + \bar{\mathbf{T}}^\top)$
\State $D_{jj} \gets \sum_k A_{jk}$
\State $\mathbf{L} \gets \mathbf{I} - \mathbf{D}^{-1/2} \mathbf{A} \mathbf{D}^{-1/2}$
\State $(\boldsymbol{\Lambda}, \mathbf{U}) \gets \mathrm{eigh}(\mathbf{L})$
\State $\lambda_\star \gets \mathrm{median}(\boldsymbol{\Lambda})$
\State Split modes into low- and high-frequency bands at $\lambda_\star$
\State $\mathbf{P}_{\mathrm{lo}} \gets \mathbf{U}_{\mathrm{lo}} \mathbf{U}_{\mathrm{lo}}^\top, \quad \mathbf{P}_{\mathrm{hi}} \gets \mathbf{U}_{\mathrm{hi}} \mathbf{U}_{\mathrm{hi}}^\top$
\State $\mathbf{A}^{\mathrm{lo}} \gets \mathbf{P}_{\mathrm{lo}} \mathbf{A} \mathbf{P}_{\mathrm{lo}}, \quad \mathbf{A}^{\mathrm{hi}} \gets \mathbf{P}_{\mathrm{hi}} \mathbf{A} \mathbf{P}_{\mathrm{hi}}$
\State $I_j \gets A^{\mathrm{lo}}_{jj}$, $\; C_j \gets \sum_{k \ne j} |A^{\mathrm{hi}}_{jk}|$
\State \textbf{return} $\mathbf{A}^{\mathrm{lo}}, \mathbf{A}^{\mathrm{hi}}, I, C, (\boldsymbol{\Lambda}, \mathbf{U})$
\end{algorithmic}
\end{algorithm}

\paragraph{Inertia--contagion correspondence.}
The inertia index of emotion $j$ is defined as the diagonal component of the low-frequency reconstruction $I_j = A^{\mathrm{lo}}_{jj}$, that is, the smooth self-loop mass after low-pass filtering. The contagion index of emotion $j$ is defined as the sum of off-diagonal components in the high-frequency reconstruction $C_j = \sum_{k \ne j} |A^{\mathrm{hi}}_{jk}|$, that is, the oscillatory inter-emotion mass.

$I_j$ is the $j$-th diagonal entry of $\mathbf{A}^{\mathrm{lo}} = \mathbf{P}_{\mathrm{lo}} \mathbf{A} \mathbf{P}_{\mathrm{lo}}$, that is, the self-loop mass of the symmetric band-limited reconstruction; $\sum_{k \ne j} |A^{\mathrm{hi}}_{jk}|$ analogously captures the off-diagonal mass of the complementary high-frequency reconstruction $\mathbf{A}^{\mathrm{hi}} = \mathbf{P}_{\mathrm{hi}} \mathbf{A} \mathbf{P}_{\mathrm{hi}}$. The empirical correspondence with the Kuppens-style emotional inertia coefficient (lag-1 autocorrelation within single-speaker subsequences~\cite{kuppens2017inertia}) is not derived analytically but is verified through the Kuppens-correspondence analysis in Appendix~\ref{app:kuppens}. The first-order Markov assumption underlying Stage 1 is examined in Appendix~\ref{app:markov}.

\subsection{Extensions and Scalability}
\label{sec:method:extensions-cost}

\paragraph{Structural extensions.}
Two extensions are introduced. First, the dialog-level hierarchical posterior replaces $\bar{\mathbf{T}}$ with a per-dialog matrix $\mathbf{T}_d$ under $\mathbf{T}_{d,j,\cdot} \mid \bar{\mathbf{T}}, \tau \sim \mathrm{Dir}(\tau \bar{\mathbf{T}}_{j,\cdot})$, where $\tau > 0$ controls the per-dialog shrinkage strength ($\tau \to \infty$ yields complete pooling, $\tau \to 0$ asymptotes to the per-dialog MLE), and the deviation $\mathrm{KL}(\mathbf{T}_d \,\|\, \bar{\mathbf{T}})$ identifies individual dialogues that deviate from the corpus norm. Second, the directed spectral decomposition adopts Chung's random-walk Laplacian $\mathbf{L}_{\mathrm{dir}} = \mathbf{I} - \tfrac{1}{2}(\mathbf{P} + \boldsymbol{\Pi}^{-1} \mathbf{P}^{\top} \boldsymbol{\Pi})$~\cite{chung2005laplacians}, where $\mathbf{P} = \bar{\mathbf{T}}$ is row-stochastic and $\boldsymbol{\Pi}$ is the diagonal Perron stationary distribution. The empirical comparison in Appendix~\ref{app:directional} confirms that the symmetrized version captures most of the structural information.

\paragraph{Computational cost.}
Stage 1 dominates the runtime. The soft-count construction is $\mathcal{O}(\sum_d |d|)$, each EB iteration is $\mathcal{O}(K^2)$ and converges within 50 iterations, and Stage 2 is the $\mathcal{O}(K^3)$ eigen-decomposition of a $K = 7$ matrix (effectively instantaneous). End-to-end runtime on a single CPU core is approximately one minute on EmotionLines ($29{,}245$ utterances), 30 seconds on MELD ($13{,}708$), and within three minutes on DailyDialog ($102{,}979$). The dialog-level cluster bootstrap ($B = 1{,}000$) is the most intensive stage but completes within 20 minutes on a single CPU core. The framework thus requires no GPU and is two to three orders of magnitude faster than Bayesian point-process models.

%% file: sections/04_experiments.tex
\section{Experiments and Evaluation}
\label{sec:experiments}

\subsection{Datasets}
\label{sec:exp:data}

BSETD is evaluated on five soft-label sources with seven-category Ekman annotations. Two are derived from real raters -- EmotionLines (5 raters) and M3ED (3 raters) -- and three are derivative sources: MELD (Dirichlet smoothing), DailyDialog (one-hot degeneration), and DailyDialog-LLM (GPT virtual annotators). EmotionLines~\cite{hsu2018emotionlines} provides $29{,}245$ utterances over $2{,}000$ dialogues (Friends TV and EmotionPush Messenger), each annotated by five independent MTurk raters; the vote distribution is preserved as a soft label. With 72.6\% of utterances exhibiting disagreement and 14.3\% lacking a majority vote, EmotionLines provides genuine human disagreement and serves as the primary evaluation target. MELD~\cite{poria2019meld} provides $13{,}708$ utterances over 1{,}433 Friends dialogues with a single hard label per utterance, from which soft labels are derived via Dirichlet smoothing (concentration $3.0$); it probes structural reproducibility under pseudo-soft labels. DailyDialog~\cite{li2017dailydialog} provides $102{,}979$ utterances over $13{,}118$ dialogues of manually written everyday conversations with hard labels, and BSETD is applied with one-hot soft labels to test structural preservation at scale. DailyDialog-LLM re-labels the same utterances with a recent large language model (GPT-5.4-mini~\cite{gptmini}) at temperature $1.0$ and $N{=}5$ independent samples under an AnnoLLM-style protocol~\cite{inoshita2026llm}; the inter-rater disagreement rate of 12.5\% is an order of magnitude lower than the 72.6\% of EmotionLines, providing a controlled probe of LLM-vs.-human disagreement. M3ED~\cite{zhao2022m3ed} provides $23{,}893$ utterances over 990 scenario dialogues from 10 Chinese television dramas, each annotated by three independent raters under the same Ekman scheme, supporting cross-lingual validation. All sources are mapped to the canonical order $(\text{neutral}, \text{joy}, \text{sadness}, \text{fear}, \text{anger}, \text{surprise}, \text{disgust})$.

\subsection{Evaluation on Emotion-Transition Pairs}
\label{sec:exp:main1}

Stage 1 of BSETD is executed on EmotionLines with EB estimation of the shared Dirichlet concentration. The posterior-mean transition matrix and the 95\% HDI of each cell are presented in Table~\ref{tab:el-top-edges}, ranked by $|\log_2 \mathrm{lift}|$. All off-diagonal entries in the BH-FDR list reject the independence null at $q = 0.10$, which reflects the large total soft-count mass ($27{,}245$ utterance pairs). Nonetheless, the effect-size ranking reveals a sharply structured pattern.

\begin{table}[t]
  \centering
  \caption{Top-10 Transition Pairs on EmotionLines, Ordered by $|\log_2 \mathrm{lift}|$. The Direction column denotes over-representation ($\uparrow$, lift $>1$) and under-representation ($\downarrow$, lift $<1$).}
  \label{tab:el-top-edges}
  \small
  \begin{tabular}{lllccc}
    \hline
    Source & Target & Direction & $\bar T_{jk}$ & lift & $\log_2$ lift \\
    \hline
    disgust & anger  & $\uparrow$ & 0.101 & 1.92 & $+0.94$ \\
    joy     & anger  & $\downarrow$ & 0.028 & 0.54 & $-0.90$ \\
    anger   & joy    & $\downarrow$ & 0.083 & 0.54 & $-0.89$ \\
    anger   & disgust & $\uparrow$ & 0.080 & 1.81 & $+0.86$ \\
    disgust & joy    & $\downarrow$ & 0.098 & 0.64 & $-0.65$ \\
    sadness & joy    & $\downarrow$ & 0.103 & 0.67 & $-0.59$ \\
    fear    & joy    & $\downarrow$ & 0.105 & 0.68 & $-0.56$ \\
    joy     & sadness & $\downarrow$ & 0.042 & 0.69 & $-0.54$ \\
    fear    & anger  & $\uparrow$ & 0.074 & 1.41 & $+0.49$ \\
    neutral & anger  & $\downarrow$ & 0.037 & 0.71 & $-0.50$ \\
    \hline
  \end{tabular}
\end{table}

The discovered structure aligns with three independent lines of psychological prediction. The pair with the largest over-representation effect, $\text{disgust} \leftrightarrow \text{anger}$ ($\log_2$ lift $+0.94$, $+0.86$), corresponds to a pair of adjacent petals on the Plutchik wheel. The pair with the largest under-representation effect, $\text{joy} \leftrightarrow \text{anger}$ ($-0.90$, $-0.89$), corresponds to a reversal along the valence axis of Russell's valence--arousal circumplex; that is, transitions between positively valenced and negatively valenced emotions are suppressed. Note that both emotions have high arousal, so the arousal axis alone cannot explain this pattern. The over-represented $\text{fear} \to \text{anger}$ transition ($+0.49$) is consistent with the appraisal-theoretic account that fear motivates aggressive defensive responses.

\subsection{Evaluation on Inertia and Contagion Separation}
\label{sec:exp:main2}

Stage 1 is fit separately on the intra-speaker (inertia) and inter-speaker (contagion) sub-matrices to test whether the same effects emerge within single-speaker turns and across speakers. Table~\ref{tab:inertia-contagion} reports the self-loop posterior means $\bar T_{jj}$ on each split together with the inertia--contagion difference.

\begin{table}[t]
  \centering
  \caption{Self-Loop Posterior Means on the Intra-Speaker (Inertia) and Inter-Speaker (Contagion) Splits of EmotionLines.}
  \label{tab:inertia-contagion}
  \small
  \begin{tabular}{lccc}
    \hline
    Emotion & Inertia $\bar T_{jj}$ & Contagion $\bar T_{jj}$ & Difference \\
    \hline
    anger    & 0.297 & 0.151 & $+0.146$ \\
    sadness  & 0.255 & 0.139 & $+0.116$ \\
    neutral  & 0.687 & 0.577 & $+0.111$ \\
    fear     & 0.165 & 0.059 & $+0.106$ \\
    disgust  & 0.158 & 0.085 & $+0.073$ \\
    surprise & 0.195 & 0.155 & $+0.039$ \\
    joy      & 0.320 & 0.298 & $+0.022$ \\
    \hline
  \end{tabular}
\end{table}

Anger and sadness exhibit the largest differences, indicating that these emotions are particularly likely to persist when the same speaker holds the floor and to dissipate when the floor passes to another interlocutor. Joy and surprise exhibit the smallest differences, indicating that their probability of being taken up by a new speaker is roughly equal to that of persisting in the original speaker, which is consistent with the high social transmissibility documented in the contagion literature of Hatfield \emph{et al.}~\cite{hatfield1993contagion}.

A more detailed effect-size analysis in Appendix~\ref{app:synthetic} shows that the over-represented $\text{disgust} \to \text{anger}$ and $\text{anger} \to \text{disgust}$ transitions are stronger on the inertia split ($\log_2$ lift $+1.28$, $+1.14$) than on the contagion split ($+0.72$, $+0.66$), supporting the hypothesis that these Plutchik-adjacent escalation patterns operate primarily within the affective state of a single speaker.

\subsection{Cross-Corpus and Cross-Lingual Evaluation}
\label{sec:exp:main3}

To assess whether the discovered structure is an artifact of a single corpus, a single language, or a single annotation source, Stage 1 is executed on all five soft-label sources under identical preprocessing, and the off-diagonal posterior means are compared pairwise. The pairwise Pearson correlations across all $42$ off-diagonal cells are reported in Table~\ref{tab:cross-corpus}.

\begin{table}[t]
  \centering
  \caption{Pairwise Off-Diagonal Pearson Correlations across the Five Sources ($n = 42$ cells). EL$=$EmotionLines, MD$=$MELD, DD$=$DailyDialog, M3$=$M3ED, LLM$=$DailyDialog-LLM.}
  \label{tab:cross-corpus}
  \footnotesize
  \setlength{\tabcolsep}{4pt}
  \begin{tabular}{lccccc}
    \hline
       & EL & MD  & DD & M3 & LLM \\
    \hline
    EL & --- & $0.910$ & $0.944$ & $0.854$ & $0.976$ \\
    MD &     & ---     & $0.916$ & $0.814$ & $0.952$ \\
    DD &     &         & ---     & $0.795$ & $0.979$ \\
    M3 &     &         &         & ---     & $0.847$ \\
    LLM &    &         &         &         & ---     \\
    \hline
  \end{tabular}
\end{table}

All six pairwise Pearson correlations are accompanied by narrow 95\% confidence intervals from the dialog-level cluster bootstrap (width $0.02$ within English and width $0.05$ across languages). The EmotionLines--MELD correlation $\rho = 0.910$ with CI $[0.899, 0.917]$ shows that structural patterns are preserved between real soft labels and pseudo-soft labels. The EmotionLines--DailyDialog correlation $0.944$ extends this preservation to a much larger corpus from a different source (manually written daily conversation), and the MELD--DailyDialog correlation $0.916$ confirms that the structure survives the transition from Dirichlet-smoothed labels to one-hot label representations. The Chinese M3ED correlates with each of the three English corpora at $0.79$--$0.85$, supporting the view that the transition structure discovered by BSETD carries a cross-lingual signal consistent with the predictions of universal emotion theory, while leaving room for language-specific and culture-specific deviations. The diagonal self-loops, in contrast, vary substantially across corpora: DailyDialog exhibits the largest neutral self-loop, and M3ED exhibits the largest negative-emotion self-loops (anger $0.65$, sadness $0.54$); see Section~\ref{sec:disc:real-vs-pseudo}.

\subsection{Evaluation on Spectral Inertia--Contagion Decomposition}
\label{sec:exp:main4}

\begin{figure*}[t]
  \centering
  \includegraphics[width=0.85\linewidth]{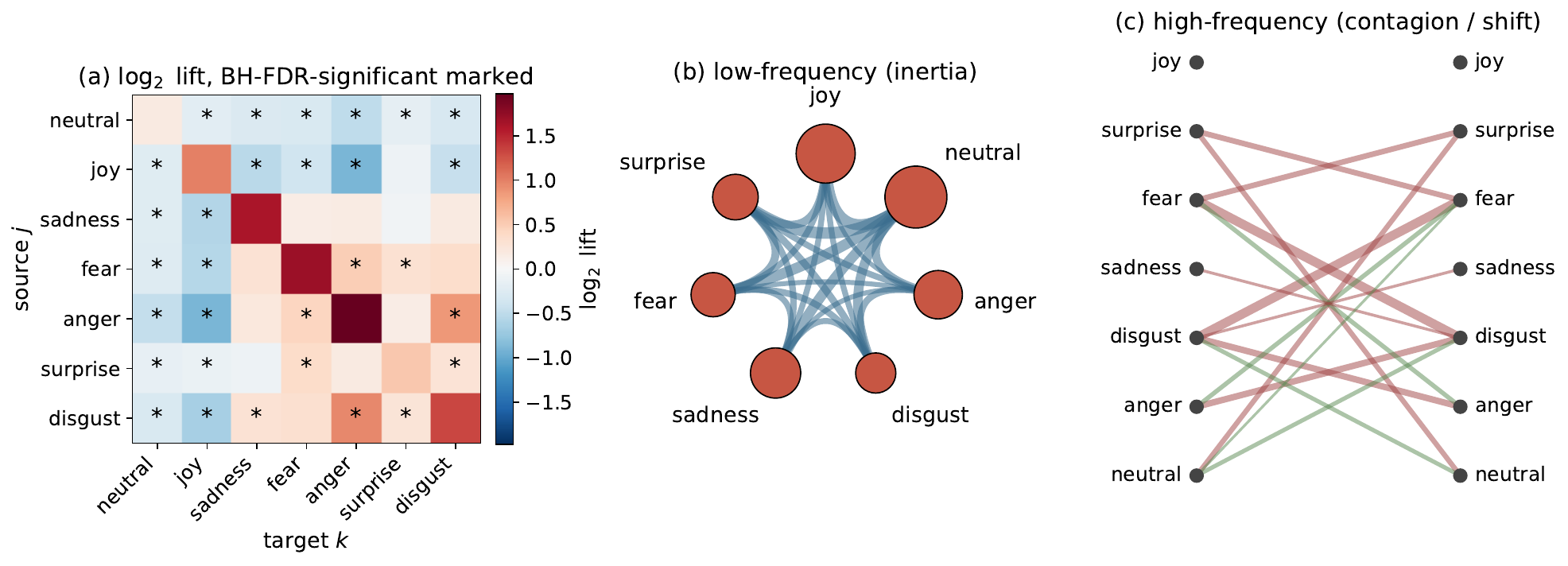}
  \caption{BSETD Results on EmotionLines. (a) $\log_2$ lift heatmap, where $*$ marks BH-FDR-significant cells. (b) Low-frequency chord diagram, with node size $=$ inertia index $I_j$ and edge thickness $= |A^{\mathrm{lo}}_{jk}|_{j \ne k}$. (c) High-frequency Sankey diagram, with ribbon width $= |A^{\mathrm{hi}}_{jk}|_{j \ne k}$.}
  \label{fig:hero}
\end{figure*}

Stage 2 is applied to the posterior-mean transition matrix on EmotionLines as the symmetrized-Laplacian spectral decomposition. The seven eigenvalues are split at the median $\lambda_\star \approx 0.88$, yielding four low-frequency modes and three high-frequency modes. Low-frequency modes correspond to smooth graph signals -- eigenvectors that vary little between adjacent nodes -- and encode near-steady-state structure, that is, the self-loop-dominated inertia component. High-frequency modes correspond to oscillatory signals that flip sign frequently between nodes, and encode the off-diagonal component, that is, the rapid cross-speaker contagion component. A three-panel display is provided in Fig.~\ref{fig:hero}.

In the heatmap (a), $\text{disgust} \to \text{anger}$ and $\text{anger} \to \text{disgust}$ appear as the deepest red and $\text{joy} \leftrightarrow \text{anger}$ as the deepest blue, directly corresponding to the Plutchik adjacency and Russell valence-axis reversal identified in main result 1. In the chord diagram (b), neutral and joy dominate in node size, visually showing that these emotions carry strong inertia in the low-frequency band. In the Sankey diagram (c), thick ribbons emanate from disgust and fear toward other emotions, showing that these emotions behave contagiously in the high-frequency band.

\begin{table}[t]
  \centering
  \caption{Stage 2 Inertia Index ($A^{\mathrm{lo}}_{jj}$) and Contagion Index ($\sum_{k\ne j} |A^{\mathrm{hi}}_{jk}|$) on EmotionLines.}
  \label{tab:spectral-indices}
  \small
  \begin{tabular}{lcc}
    \hline
    Emotion & Inertia index & Contagion index \\
    \hline
    neutral  & 0.378 & 0.025 \\
    joy      & 0.338 & 0.011 \\
    sadness  & 0.227 & 0.010 \\
    fear     & 0.162 & 0.046 \\
    anger    & 0.206 & 0.028 \\
    surprise & 0.177 & 0.031 \\
    disgust  & 0.123 & 0.052 \\
    \hline
  \end{tabular}
\end{table}

Table~\ref{tab:spectral-indices} reports the inertia and contagion indices for each emotion. Neutral and joy dominate the inertia ranking (inertia $0.378$, $0.338$), while disgust and fear dominate the contagion ranking (contagion $0.052$, $0.046$). This contrast is consistent with two psychological interpretations. First, it agrees with Hatfield's emotional contagion theory, which holds that high-arousal negative emotions (disgust, fear, anger) are readily transmitted across speakers. Second, it agrees with Kuppens' emotion-dynamics theory, which holds that neutral and positively valenced joy persist inertially within a single speaker. More importantly, the inertia--contagion correspondence is not assumed but arises naturally from the data-driven Laplacian spectral decomposition: a single operation, median splitting at $\lambda_\star$, automatically achieves the separation of components corresponding to these two psychological constructs.

\subsection{Evaluation on Robustness}
\label{sec:exp:ablation}

The robustness of the BSETD claims is verified along five aspects (uncertainty, null distribution, subpopulation, model selection, downstream prediction) in Table~\ref{tab:ablation}. The 95\% confidence intervals of the inertia--contagion differences, constructed by dialog-level cluster bootstrap ($B = 1{,}000$), do not contain zero for any of the seven emotions (e.g., joy $[+0.003, +0.042]$, anger $[+0.112, +0.177]$). Furthermore, the within-dialog permutation null test ($B = 50$) shows that the empirical $\max |\log_2 \mathrm{lift}| = 0.94$ exceeds all permutations ($p < 0.02$), demonstrating that the directional structure does not arise from the marginal distribution alone.

Across six subpopulations of EmotionLines partitioned by dialog attributes, pairwise Pearson correlations range over $[0.873, 0.978]$; under rater subsampling with $R \in \{2, 3, 4\}$, the Pearson correlation to the $R=5$ baseline is $0.991$--$0.999$; and the cross-domain correlation between Friends and EmotionPush is $\rho = 0.962$. The structure is thus robust to dialog length, disagreement level, number of speakers, rater budget, and media type. On the model-selection axis, the lift ranking is unchanged under hard-labelization, under the fixed prior $\boldsymbol{\alpha} = \mathbf{1}$, and under equal-width spectral splitting; and replacing the symmetrized Laplacian with the Chung directed Laplacian (Appendix~\ref{app:directional}) yields Pearson correlations of $0.91 / 0.80$ for the inertia and contagion indices. On the downstream-prediction axis, adding BSETD lift features yields macro-F1 of $0.265$ (unchanged) for a linear logistic regression (lag-5), $0.191 \to 0.205$ ($+0.015$) for a 2-stream GRU, and $0.434 \to 0.440$ ($+0.007$) for a frozen DistilBERT (lag-5). The contribution shrinks as the encoder capacity grows, which supports the positioning of BSETD as a framework whose primary value lies in structural discovery rather than utterance-level prediction.

\begin{table*}[t]
  \centering
  \caption{Robustness-Ablation Summary on EmotionLines.}
  \label{tab:ablation}
  \footnotesize
  \setlength{\tabcolsep}{4pt}
  \begin{tabular}{lcc}
    \hline
    Ablation & Metric & Result \\
    \hline
    Dialog cluster bootstrap & Lower bound of $I-C$ 95\% CI & joy $+0.003$, anger $+0.112$ \\
    Within-dialog permutation & $p$-value ($B=50$) & $<0.02$ \\
    Subpopulation (6 splits) & Pairwise Pearson range & $[0.873, 0.978]$ \\
    Rater $R = 2$ subsample & Pearson vs.\ $R=5$ & $0.991$ \\
    Rater $R = 3$ subsample & Pearson vs.\ $R=5$ & $0.997$ \\
    Rater $R = 4$ subsample & Pearson vs.\ $R=5$ & $0.999$ \\
    Friends vs.\ EmotionPush & Pairwise Pearson & $0.962$ \\
    Hard-label substitution & Posterior-mean MAD (top-disagreement decile) & $0.014$ \\
    Fixed prior $\alpha=\mathbf{1}$ & Lift-ranking change & none \\
    Directed Laplacian (Chung) & Inertia/contagion Pearson & $0.91 / 0.80$ \\
    \hline
    LogReg + lag-5 & BSETD-lift macro-F1 & $0.265$ (no gain) \\
    GRU 2-stream & BSETD-lift macro-F1 & $0.191 \to 0.205$ \\
    DistilBERT (frozen) + lag-5 & BSETD-lift macro-F1 & $0.434 \to 0.440$ \\
    \hline
  \end{tabular}
\end{table*}

%% file: sections/05_discussion.tex
\section{Discussion}
\label{sec:discussion}

\subsection{Psychological Meaning of the Discovered Edges}
\label{sec:disc:psych}

The empirical structure recovered by BSETD on EmotionLines aligns with three independent lines of affective science.

\paragraph{Plutchik adjacency and Russell valence-axis reversal.}
The pair with the largest over-representation effect, $\text{disgust} \to \text{anger}$ ($\log_2$ lift $+0.94$) and $\text{anger} \to \text{disgust}$ ($+0.86$), corresponds directly to a pair of adjacent petals on the Plutchik wheel; the symmetric shape of both directions being over-represented at nearly the same magnitude is exactly what would be expected from transitions between two points close in affective space. The pair with the largest under-representation effect, $\text{joy} \to \text{anger}$ ($-0.90$) and $\text{anger} \to \text{joy}$ ($-0.89$), does not lie opposite on the Plutchik wheel (anticipation is interposed between them) but corresponds to a reversal along the valence axis in Russell's valence--arousal model; since both emotions are high-arousal, the pattern is explicable only through valence. Thus, without any distance supervision, the signatures of two distinct affective spaces are simultaneously recovered from the transition statistics alone.

\paragraph{Gottman's escalation cascade.}
Gottman's longitudinal studies on dyadic conflict identified $\text{anger} \to \text{contempt/disgust}$ as a robust escalation pattern that predicts relationship rupture. Our inertia-split analysis shows that this same $\text{anger} \to \text{disgust}$ transition concentrates in intra-speaker turns (inertia $\log_2$ lift $+1.14$) more strongly than in inter-speaker turns (contagion $\log_2$ lift $+0.66$), which is consistent with the clinical interpretation that the escalation is driven by the deepening of a speaker's own appraisal rather than by a cycle initiated by the partner.

\paragraph{Hatfield's emotional contagion.}
The over-represented $\text{fear} \to \text{anger}$ transition ($\log_2$ lift $+0.49$) is the only inter-emotion pair for which the lift on the contagion split exceeds that on the inertia split. It shows that fear is more likely to evoke anger in another speaker than to persist as anger in the same speaker -- precisely the asymmetric pattern that Hatfield's contagion framework predicts for high-arousal negative emotions in dyadic interaction.

\subsection{Why Both Stages Are Necessary}
\label{sec:disc:both-stages}

Stage 1 alone yields the posterior over the transition matrix and an effect-size ranking, but it cannot separate the two qualitatively distinct sources of structure: smooth self-loop-dominated mass and oscillatory off-diagonal mass. Stage 2 makes this distinction explicit by projecting onto a frequency-ordered basis derived from the graph itself. The empirical result on EmotionLines is that neutral and joy dominate the low-frequency band (inertia indices $0.378$, $0.338$), while disgust and fear dominate the high-frequency band (contagion indices $0.052$, $0.046$). This separation is not assumed but emerges naturally from the spectral decomposition of the data-driven Laplacian.

\subsection{Real vs.\ Pseudo Soft Labels}
\label{sec:disc:real-vs-pseudo}

For six of the seven emotions, the diagonal self-loops on EmotionLines exceed those on MELD, with a mean difference of $+0.079$. The most natural explanation is that the Dirichlet-smoothed pseudo-soft labels on MELD homogenize the within-utterance distribution and pull mass away from the modal category, thereby weakening the apparent diagonal of the transition matrix. The real five-rater soft labels of EmotionLines preserve a bimodal disagreement structure (either a single rater wins decisively or the votes split into two clusters), and this translates into sharp self-loop posteriors at the corpus level. The single emotion that bucks this trend, fear ($-0.072$), is the rarest emotion in EmotionLines ($P(\text{fear}) = 0.029$ marginally) and appears at a substantially higher base rate in the Friends sample of MELD. While the MELD posterior is supported by more soft counts, the EmotionLines posterior shrinks toward the EB prior; the fear gap is therefore attributed to a power-related artifact rather than a structural reversal. The off-diagonal pattern is nonetheless preserved at Pearson $0.91$ (Section~\ref{sec:exp:main3}), implying that the qualitative structure -- which transitions exist -- is robust to the soft-label generation process, while the magnitude of inertia for high-base-rate categories is sensitive to it.

\subsection{Cross-Lingual Robustness and Downstream Utility}
\label{sec:disc:cross-lingual-downstream}

The cross-lingual correlation with M3ED ($0.79$--$0.85$) lies below the within-English range ($0.91$--$0.98$), but it remains consistent with a universal pattern of basic-category emotion dynamics and is difficult to explain by purely language-specific artifacts. M3ED exhibits larger anger and sadness self-loops than any of the English sources, which is thought to reflect the cultural conventions of emotional expression in scripted Chinese television. The DailyDialog-LLM comparison is informative in a different direction: replacing the one-hot hard labels on the same utterance set with GPT-5.4-mini $N{=}5$ virtual votes yields a Pearson correlation of $0.979$ on the BSETD off-diagonal structure, indicating that structural discovery is essentially invariant to the source of the soft labels. At the same time, the inter-rater disagreement rate of the LLM-soft labels is 12.5\%, an order of magnitude lower than the 72.6\% of the real five raters in EmotionLines, which is consistent with the AnnoLLM-style observation~\cite{inoshita2026llm} that LLMs reproduce modal-label structure while strongly compressing human uncertainty distributions. The recovered structural agreement should therefore be interpreted as agreement on modal-transition patterns, not as a substitute for genuine multi-rater human signal. In downstream prediction, the contribution of BSETD lift features shrinks with the representational capacity of the downstream encoder (Section~\ref{sec:exp:ablation}), reinforcing the positioning of the framework as one whose primary value lies in structural discovery rather than utterance-level prediction.

\subsection{Limitations}
\label{sec:disc:limits}

Three limitations remain. (i) Because the analysis aggregates transitions over the entire corpus, intra-dialog heterogeneity -- such as a dialogue whose first half is dominated by joy--trust transitions while its second half is dominated by anger--disgust transitions -- appears as a mixed signal. The dialog-level hierarchical extension in Section~\ref{sec:method:extensions-cost} is the natural remedy. (ii) The symmetrized decomposition of Stage 2 discards directional information; while the speaker-conditioned sub-matrices and the Chung directed Laplacian (Appendix~\ref{app:directional}) partially compensate for this loss, a fully directional analysis is left for future work. (iii) The outer-product formulation assumes conditional independence between adjacent labels, which may be violated under strongly correlated annotator behavior.

%% file: sections/06_conclusion.tex
\section{Conclusion}
\label{sec:conclusion}

In this article, we have introduced BSETD, a two-stage framework that combines hierarchical DM posterior estimation with the spectral decomposition of a symmetrized graph Laplacian. The Bayesian stage propagates annotator uncertainty through a closed-form outer-product formulation, and the spectral stage separates low-frequency inertia from high-frequency contagion, providing a principled bridge to Kuppens-style inertia and Hatfield-style contagion. On EmotionLines, the Plutchik-adjacent pair disgust$\to$anger and anger$\to$disgust is over-represented at $\log_2$ lift $+0.94$ and $+0.86$, the Russell-valence-reversed pair joy$\to$anger and anger$\to$joy is under-represented at $-0.90$ and $-0.89$, and the inertia--contagion difference is positive for all seven emotions, with the largest values observed for anger ($+0.146$) and sadness ($+0.116$). A five-source validation yields pairwise Pearson correlations of $0.91$--$0.98$ within English, $0.79$--$0.85$ against Chinese M3ED, and $0.979$ between human hard labels and LLM virtual soft labels on the same utterances. 

In future studies, we will investigate per-dialog hierarchical posteriors that expose intra-dialog heterogeneity, a fully directional spectral analysis, and the extent to which virtual-annotator LLM pipelines can partially substitute for human raters, in light of the known modal-vs.-uncertainty gap. 

%% file: sections/A_appendix.tex
\appendix

\section{Identifiability and Sample Complexity}
\label{app:theory}

\subsection{Identifiability of $\mathbf{T}$}
\label{app:identifiability}

The soft-label outer product of Eq.~\eqref{eq:soft-counts} is identifiable for $\mathbf{T}$ under two assumptions: (i) $\mathbf{p}_{d,t}$ is an unbiased estimator of the utterance-level posterior, $\mathbb{E}[p_{d,t}(j)] = \Pr(e_{d,t} = j \mid u_{d,t})$; and (ii) $\{e_{d,t}\}$ is a Markov chain with transition $\mathbf{T}$ conditionally independent of the soft-label noise. Under these assumptions, $\mathbb{E}[N_{jk}] = \sum_{d,t} \Pr(e_{d,t} = j) \, T_{jk}$. Letting $S_j = \sum_{d,t} \Pr(e_{d,t} = j)$, the relation $T_{jk} = \mathbb{E}[N_{jk}] / S_j$ holds, and $\mathbf{T}$ is identifiable from finite-sample $\mathbf{N}$ whenever $S_j > 0$ for all $j$. When (i) is violated, the estimate corresponds to the soft-label-conditional transition; even then, the lift remains a valid indicator of association.

\subsection{Sample Complexity of Lift Recovery}
\label{app:sample-complexity}

Standard DM sample-complexity arguments give $\mathrm{Var}[T_{jk} \mid \mathbf{N}] \le \bar T_{jk}(1 - \bar T_{jk}) / (S_j + \sum_k \alpha^\star_k + 1)$. To detect a true lift $r = T_{jk} / \bar P_k$ with relative error $\epsilon$ under 95\% coverage, the source row requires $S_j \gtrsim (\bar P_k \cdot r) / (\epsilon^2 \bar T_{jk})$. For $S_{\text{joy}} = 4{,}193$ on EmotionLines and $\epsilon = 0.10$, this is comfortably satisfied above the empirical noise floor.

\subsection{Granger-Causal Interpretation}
\label{app:granger}

A $j \to k$ edge asserts a corpus-level conditional association $\Pr(e_{t+1} = k \mid e_t = j) > \Pr(e_{t+1} = k)$, that is, the population analogue of discrete-state Granger predictability. BSETD does not test interventional causality and cannot distinguish actively transmitted emotion from a co-occurring common cause.

\section{Empirical Validation of the Markov Assumption}
\label{app:markov}

Stage 1 relies on first-order Markov conditional independence. On EmotionLines, the empirical two-step transition matrix $P_2$, estimated from second-neighbor pairs via the outer-product formulation, is compared with the Markov prediction $T^2 = \bar{\mathbf{T}} \bar{\mathbf{T}}$. The mean per-source KL divergence $\mathrm{KL}(P_{2,j} \,\|\, T^2_{j})$ is $0.039$ (max $0.093$ for anger), versus an independence baseline of $0.067$, giving a 42\% improvement. The Markov assumption is therefore a defensible first-order approximation, and the largest deviation concentrating on anger is consistent with persistent anger in dyadic conflict, suggesting that higher-order extensions are a natural direction for future work.

\section{Directed Spectral Decomposition}
\label{app:directional}

Two spectral decompositions are compared on the BSETD posterior-mean transition matrix on EmotionLines. The first is the symmetrized normalized Laplacian used in the main text. The second is the Chung directed random-walk Laplacian (Section~\ref{sec:method:extensions-cost}), with the Perron stationary distribution $\boldsymbol{\pi}$ used as the diagonal weights. The Pearson correlations between the inertia and contagion indices recovered by the two decompositions are $0.91$ and $0.80$, respectively. The symmetrized version captures most of the structural information, while the directed refinement adds modest additional signal to the contagion component.

\section{Empirical Correspondence with the Kuppens Inertia Coefficient}
\label{app:kuppens}

The inertia captured by BSETD is empirically validated on EmotionLines against the Kuppens emotional-inertia coefficient~\cite{kuppens2017inertia} (lag-1 autocorrelation within single-speaker subsequences). For each emotion $j$, the lag-1 Pearson autocorrelation $\rho^{\mathrm{Kup}}_j$ of the soft probability $p_{d,t}(j)$ within adjacent same-speaker turns is computed, yielding anger $+0.58$, neutral $+0.48$, sadness $+0.44$, fear $+0.43$, joy $+0.41$, disgust $+0.39$, surprise $+0.36$ (all $p < 10^{-3}$). The correspondence with the Stage 1 inertia--contagion difference $\bar T^{\mathrm{in}}_{jj} - \bar T^{\mathrm{co}}_{jj}$ is strong (Pearson $+0.80$, $p = 0.029$; Spearman $+0.86$, $p = 0.014$), because both quantities measure within-speaker persistence. The weaker correspondence with $I_j$ alone (Pearson $+0.25$) reflects that $I_j$ is obtained through graph-spectral low-frequency reconstruction rather than speaker conditioning; the two should be interpreted as complementary.

\section{Synthetic Validation of Stage 1}
\label{app:synthetic}

Stage 1 recovery is evaluated on a synthetic dialogue corpus generated from four ground-truth topologies (chain, fork, star, cycle) over a grid of 144 settings: number of dialogues $N \in \{200, 500\}$, average length $L \in \{10, 20\}$, rater accuracy $p_{\text{acc}} \in \{0.55, 0.75, 0.95\}$, and three random seeds. Lift-ranking recovery is consistently stronger than BH-FDR-based recovery: the top-$|E|$ lift recall is $1.0$ for chain and cycle, $0.83$--$1.0$ for star, and $0.33$--$0.55$ for fork. The fork topology, in which a single source distributes mass to many targets, is the most difficult regime for corpus-level structural recovery, and is reported honestly as a known failure mode.

%% file: references_bsetd.bib
@inproceedings{dialoguernn2019,
  author    = {Navonil Majumder and Soujanya Poria and Devamanyu Hazarika and Rada Mihalcea and Alexander Gelbukh and Erik Cambria},
  title     = {DialogueRNN: An Attentive RNN for Emotion Detection in Conversations},
  booktitle = {Proceedings of the AAAI Conference on Artificial Intelligence},
  year      = {2019},
  number = {837},
  pages = {6818--6825},
  doi       = {10.1609/aaai.v33i01.33016818},
}

@book{picard1997affective,
  author    = {Picard, Rosalind W.},
  title     = {Affective Computing},
  publisher = {MIT Press},
  year      = {1997},
  doi = {10.7551/mitpress/1140.001.0001}
}

@article{emotionIC2024,
  author    = {Yingjian Liu and Jiang Li and Xiaoping Wang and Zhigang Zeng},
  title     = {{EmotionIC}: Emotional Inertia and Contagion-driven Dependency Modeling for Emotion Recognition in Conversation},
  journal   = {Science China Information Sciences},
  year      = {2024},
  volume = {67},
  number = {182103},
  doi      = {10.1007/s11432-023-3908-6}
}

@article{graphsmile2023,
  author    = {Jiang Li and Xiaoping Wang and Zhigang Zeng},
  title     = {Tracing Intricate Cues in Dialogue: Joint Graph Structure and Sentiment Dynamics for Multimodal Emotion Recognition},
  journal   = {IEEE Transactions on Pattern Analysis and Machine Intelligence},
  year      = {2025},
  pages = {8786--8803},
  doi    = {10.1109/TPAMI.2025.3581236},
}

@inproceedings{dgode2024,
  author    = {Yuntao Shou and Tao Meng and Wei Ai and Keqin Li},
  title     = {Dynamic Graph Neural ODE Network for Multi-modal Emotion Recognition in Conversation},
  booktitle = {Proceedings of the International Conference on Computational Linguistics},
  year      = {2025},
  pages = {256--268},
}

@inproceedings{hsu2018emotionlines,
  author    = {Chao-Chun Hsu and Sheng-Yeh Chen and Chuan-Chun Kuo and Ting-Hao Huang and Lun-Wei Ku},
  title     = {EmotionLines: An Emotion Corpus of Multi-Party Conversations},
  booktitle = {Proceedings of the Eleventh International Conference on Language Resources and Evaluation},
  year      = {2018},
}

@inproceedings{lewidi2025,
  author    = {Elisa Leonardelli and Silvia Casola and Siyao Peng and Giulia Rizzi and Valerio Basile and Elisabetta Fersini and Diego Frassinelli and Hyewon Jang and Maja Pavlovic and Barbara Plank and Massimo Poesio},
  title     = {LeWiDi-2025 at NLPerspectives: Third Edition of the Learning with Disagreements Shared Task},
  booktitle = {Proceedings of the The 4th Workshop on Perspectivist Approaches to NLP},
  year      = {2025},
  pages = {182--195},
  doi      = {10.18653/v1/2025.nlperspectives-1.16}
}

@inproceedings{perspectivist2021,
  author    = {Federico Cabitza and Andrea Campagner and Valerio Basile},
  title     = {Toward a Perspectivist Turn in Ground Truthing for Predictive Computing},
  booktitle   = {Proceedings of the Thirty-Seventh AAAI Conference on Artificial Intelligence and Thirty-Fifth Conference on Innovative Applications of Artificial Intelligence and Thirteenth Symposium on Educational Advances in Artificial Intelligence},
  year      = {2023},
  number = {771},
  pages = {6860--6868},
  doi      = {10.1609/aaai.v37i6.25840}
}

@misc{inoshita2026llm,
  author    = {Keito Inoshita and Xiaokang Zhou and Akira Kawai and  Katsutoshi Yada},
  title     = {LLMs Capture Emotion Labels, Not Emotion Uncertainty: Distributional Analysis and Calibration of Human-LLM Judgment Gaps},
  year      = {2026},
  archivePrefix = {arXiv},
  doi      = {10.48550/arXiv.2604.27345}
}

@inproceedings{wang2025intentdrift,
  author    = {Yue Wang and Dehang Fu and Jie Tan and Junxiao Han and Yao Wan and Lixin Cui and Lu Bai and Philip S. Yu},
  title     = {Detecting Intent Drift in Continuous Conversation via Temporal Transition Accumulation},
  booktitle = {Proceedings of the IEEE International Conference on Data Mining},
  year      = {2025},
  pages = {773-782},
  doi      = {10.1109/ICDM65498.2025.00085}
}

@inproceedings{mecot2025,
  author    = {Yangbo Wei and Zhen Huang and Fangzhou Zhao and Qi Feng and Wei W. Xing},
  title     = {{MECoT}: Markov Emotional Chain-of-Thought for Personality-Consistent Role-Playing},
  booktitle = {Findings of the Association for Computational Linguistics: ACL},
  year      = {2025},
  pages     = {8297--8314},
  note      = {10.18653/v1/2025.findings-acl.435}
}

@inproceedings{kbcin_aaai2023,
  author    = {Weixiang Zhao and Yanyan Zhao and Zhuojun Li and Bing Qin},
  title     = {Knowledge-Bridged Causal Interaction Network for Causal Emotion Entailment},
  booktitle = {Proceedings of the Thirty-Seventh AAAI Conference on Artificial Intelligence and Thirty-Fifth Conference on Innovative Applications of Artificial Intelligence and Thirteenth Symposium on Educational Advances in Artificial Intelligence},
  year      = {2023},
  number = {1572},
  pages = {14020--14028},
  doi      = {10.1609/aaai.v37i11.2664}
}

@inproceedings{gssmcc_aaai2025,
  author    = {Wei Ai and Fuchen Zhang and Yuntao Shou and Tao Meng and Haowen Chen and Keqin Li},
  title     = {Revisiting Multimodal Emotion Recognition in Conversation from the Perspective of Graph Spectrum},
  booktitle = {Proceedings of the Thirty-Ninth AAAI Conference on Artificial Intelligence and Thirty-Seventh Conference on Innovative Applications of Artificial Intelligence and Fifteenth Symposium on Educational Advances in Artificial Intelligence},
  year      = {2025},
  number = {1269},
  pages = {11418--11426},
  doi    = {10.1609/aaai.v39i11.3324},
}

@article{chung2005laplacians,
  author    = {Fan Chung},
  title     = {Laplacians and the Cheeger Inequality for Directed Graphs},
  journal   = {Annals of Combinatorics},
  year      = {2005},
  volume    = {9},
  pages     = {1--19},
  doi       = {10.1007/s00026-005-0237-z},
}

@article{cowen2017emotion,
  author    = {Alan S. Cowen and Dacher Keltner},
  title     = {Self-report captures 27 distinct categories of emotion bridged by continuous gradients},
  journal   = {Proceedings of the National Academy of Sciences},
  year      = {2017},
  volume    = {114},
  number    = {38},
  pages     = {E7900--E7909},
  doi       = {10.1073/pnas.1702247114}
}

@article{kuppens2017inertia,
  author    = {Peter Kuppens and Philippe Verduyn},
  title     = {Emotion dynamics},
  journal   = {Current Opinion in Psychology},
  year      = {2017},
  volume    = {17},
  pages     = {22--26},
  doi       = {10.1016/j.copsyc.2017.06.004}
}

@article{hatfield1993contagion,
  author    = {Elaine Hatfield and John T. Cacioppo and Richard L. Rapson},
  title     = {Emotional Contagion},
  journal   = {Current Directions in Psychological Science},
  year      = {1993},
  volume    = {2},
  number    = {3},
  pages     = {96--99},
  doi       = {10.1111/1467-8721.ep10770953},
}

@article{loossens2020affective,
  author    = {Tim Loossens and Merijn Mestdagh and Egon Dejonckheere and Peter Kuppens and Francis Tuerlinckx and Stijn Verdonck},
  title     = {The Affective Ising Model: A computational account of human affect dynamics},
  journal   = {PLOS Computational Biology},
  year      = {2020},
  volume    = {16},
  number    = {5},
  pages     = {e1007860},
  doi       = {10.1371/journal.pcbi.1007860},
}

@techreport{minka2000dirichlet,
  author    = {Thomas P. Minka},
  title     = {Estimating a {D}irichlet Distribution},
  institution = {Microsoft},
  year      = {2000},
  pages = {1--15}
}

@article{shuman2013emerging,
  author    = {David I. Shuman and Sunil K. Narang and Pascal Frossard and Antonio Ortega and Pierre Vandergheynst},
  title     = {The emerging field of signal processing on graphs: Extending high-dimensional data analysis to networks and other irregular domains},
  journal   = {IEEE Signal Processing Magazine},
  year      = {2013},
  volume    = {30},
  number    = {3},
  pages     = {83--98},
  doi       = {10.1109/MSP.2012.2235192},
}

@inproceedings{poria2019meld,
  author    = {Soujanya Poria and Devamanyu Hazarika and Navonil Majumder and Gautam Naik and Erik Cambria and Rada Mihalcea},
  title     = {{MELD}: A Multimodal Multi-Party Dataset for Emotion Recognition in Conversations},
  booktitle = {Proceedings of the Annual Meeting of the Association for Computational Linguistics},
  year      = {2019},
  pages = {527-536},
  doi      = {10.18653/v1/P19-1050}
}

@inproceedings{li2017dailydialog,
  author    = {Yanran Li and Hui Su and Xiaoyu Shen and Wenjie Li and Ziqiang Cao and Shuzi Niu},
  title     = {{DailyDialog}: A Manually Labelled Multi-turn Dialogue Dataset},
  booktitle = {Proceedings of the International Joint Conference on Natural Language Processing},
  year      = {2017},
  pages = {986--995}
}

@inproceedings{zhao2022m3ed,
  author    = {Jinming Zhao and Tenggan Zhang and Jingwen Hu and Yuchen Liu and Qin Jin and Xinchao Wang and Haizhou Li},
  title     = {{M3ED}: Multi-modal Multi-scene Multi-label Emotional Dialogue Database},
  booktitle = {Proceedings of the Annual Meeting of the Association for Computational Linguistics},
  year      = {2022},
  pages = {5699--5710},
  doi      = {10.18653/v1/2022.acl-long.391}
}

@article{benjamini1995fdr,
  author    = {Yoav Benjamini and Yosef Hochberg},
  title     = {Controlling the False Discovery Rate: A Practical and Powerful Approach to Multiple Testing},
  journal   = {Journal of the Royal Statistical Society, Series B},
  year      = {1995},
  volume    = {57},
  number    = {1},
  pages     = {289--300},
  doi       = {10.1111/j.2517-6161.1995.tb02031.x},
}

@article{iemocap2008,
  author    = {Carlos Busso and Murtaza Bulut and Chi-Chun Lee and Abe Kazemzadeh and Emily Mower and Samuel Kim and Jeannette N. Chang, Sungbok Lee and Shrikanth S. Narayanan},
  title     = {IEMOCAP: interactive emotional dyadic motion capture database},
  journal   = {Language Resources and Evaluation},
  year      = {2008},
  volume    = {42},
  pages     = {335--359},
  doi       = {10.1007/s10579-008-9076-6},
}

@inproceedings{gptmini,
    author = {OpenAI},
    title = {GPT-5.4 mini},
    booktitle = {OpenAI Developers},
    year = {2026},
    note = {https://developers.openai.com/api/docs/models/gpt-5.4-mini}
}
